\documentclass[letterpaper, 10 pt, conference]{ieeeconf}  

\IEEEoverridecommandlockouts                              

\usepackage{bbm}
\usepackage{etoolbox}
\usepackage{multicol}
\usepackage{cite}
\usepackage[bookmarks=true]{hyperref}
\usepackage{graphics} 
\usepackage{times} 
\usepackage{amsmath} 
\usepackage{amssymb}  
\usepackage{mathrsfs}
\usepackage{amsfonts}
\usepackage{wrapfig}
\usepackage{amsmath}

\usepackage{kantlipsum}
\usepackage{subcaption}
\usepackage{marvosym}

\usepackage{graphicx}
\usepackage{float}
\usepackage{ctable}
\usepackage{cuted}
\usepackage{colortbl}
\usepackage{multirow}
\usepackage[misc,geometry]{ifsym}
\usepackage{algorithm}
\usepackage{algpseudocode}
\usepackage{xcolor}   
\usepackage{pifont}    
\usepackage{pifont}
\usepackage{array}
\usepackage{tabularx}
\usepackage[table]{xcolor}
\usepackage{amssymb}
\usepackage{multicol}
\usepackage{adjustbox}
\let\labelindent\relax
\usepackage{enumitem}
\usepackage{makecell} 

\makeatletter
\let\NAT@parse\undefined
\makeatother
\usepackage{hyperref}

\newcommand{\cmark}{\textcolor{green}{\checkmark}}
\newcommand{\xmark}{\textcolor{red}{\sffamily X}}

\definecolor{ImportantColor}{rgb}{0.85, 0.85, 0.85}

\title{\LARGE \bf
RoTri-Diff: A Spatial Robot–Object Triadic Interaction-Guided Diffusion Model for Bimanual Manipulation 
}

\author{\normalsize Zixuan Chen$^{1,3}$, Nga Teng Chan$^{4,3}$, Yiwen Hou$^{3}$, Chenrui Tie$^{3}$, Zixuan Liu$^{3}$, Haonan Chen$^{3}$, Junting Chen$^{3}$,\\ Jieqi Shi$^{2}$, Yang Gao$^{2}$, Jing Huo$^{1\dag}$, Lin Shao$^{3\dag}$
\thanks{\textbf{$^{\dagger}$Corresponding Author}}
\thanks{This work was completed during Zixuan and Nga Teng's visiting at the National University of Singapore.}
\thanks{$^{1}$Zixuan Chen and Jing Huo are with the School of Computer Science, Nanjing University, China. Emails: {\texttt{chenzx@nju.edu.cn, huojing@nju.edu.cn}}}
\thanks{$^{2}$Jieqi Shi and Yang Gao are with the School of Intelligence Science and Technology, Nanjing University, China. Emails: {\texttt{isjieqi@nju.edu.cn, gaoy@nju.edu.cn}}}
\thanks{$^{3}$Yiwen Hou, Chenrui Tie, Zixuan Liu, Haonan Chen, Junting Chen and Lin Shao are with the School of Computing, National University of Singapore, Singapore. Email: {\texttt{\{yiwenhou,chenrui.tie,zixuanliu\}@u.nus.edu, \{chenhaonan,junting.chen,linshao\}@u.nus.edu}}}
\thanks{$^{4}$Nga Teng Chan is with the Department of Computer Science and Engineering, The Hong Kong University of Science and Technology, China. Email: {\texttt{ntchanab@connect.ust.hk}}}
\thanks{This work is supported in part by New Generation Artificial Intelligence-National Science and Technology Major Project (2025ZD0122904), National Natural Science Foundation of China (62192783, 62276128, 62506153), Jiangsu Science and Technology Major Project (BG2025035), the Fundamental Research Funds for the Central Universities(KG202514) and the Collaborative Innovation Center of Novel Software Technology and Industrialization.}
}

\begin{document}


\maketitle

\begin{strip}
\vspace{-57mm} 
\begin{center}
    \includegraphics[width=0.9\textwidth]{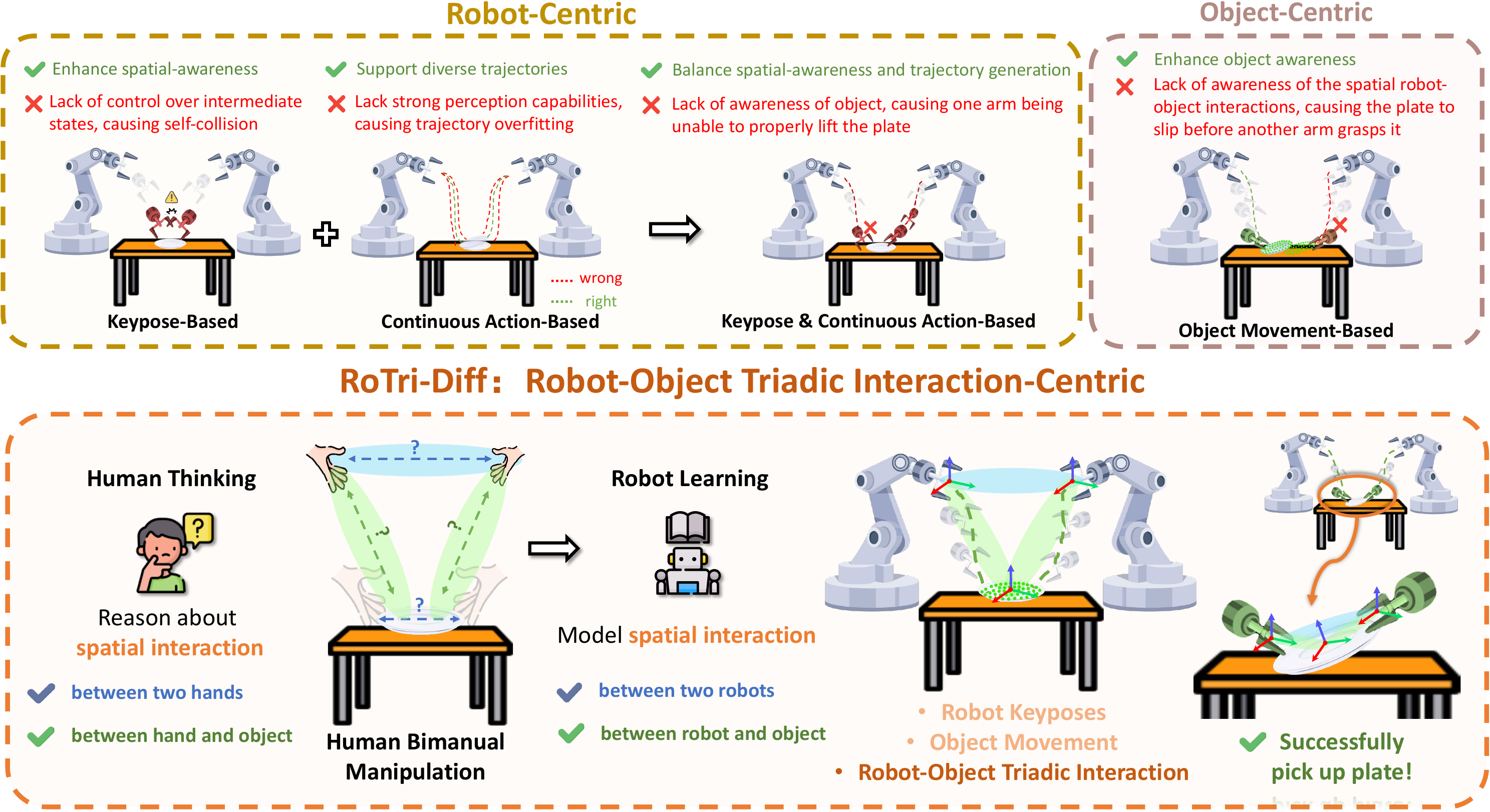}
    \vspace{-0.8mm}
    \captionof{figure}{We present \textbf{RoTri-Diff}, a diffusion-based framework for bimanual imitation learning that centers on robot–object triadic interaction (\textit{RoTri}). By explicitly modeling and leveraging the relative 6D pose relations between the two arm end-effectors and the manipulated objects, it achieves stable performance on bimanual tasks requiring fine-grained coordination.}
    \label{fig:teaser}
\end{center}
\vspace{-5mm}
\end{strip}

\thispagestyle{empty}
\pagestyle{empty}

\begin{abstract}
Bimanual manipulation is a fundamental robotic skill that requires continuous and precise coordination between two arms. While imitation learning (IL) is the dominant paradigm for acquiring this capability, existing approaches, whether robot-centric or object-centric, often overlook the dynamic geometric relationship among the two arms and the manipulated object. This limitation frequently leads to inter-arm collisions, unstable grasps, and degraded performance in complex tasks.
To address this, in this paper we explicitly models the Robot–Object Triadic Interaction (\textit{RoTri}) representation in bimanual systems, by encoding the relative 6D poses between the two arms and the object to capture their spatial triadic relationship and establish continuous triangular geometric constraints.
Building on this, we further introduce RoTri-Diff, a diffusion-based imitation learning framework that combines \textit{RoTri} constraints with robot keyposes and object motion in a hierarchical diffusion process. This enables the generation of stable, coordinated trajectories and robust execution across different modes of bimanual manipulation.
Extensive experiments show that our approach outperforms state-of-the-art baselines by 10.2\% on 11 representative RLBench2 tasks and achieves stable performance on 4 challenging real-world bimanual tasks.
Project website: \href{https://rotri-diff.github.io/}{https://rotri-diff.github.io/}.
\end{abstract}
\section{Introduction}
Bimanual manipulation is a fundamental robotic capability, essential for executing complex, human-like tasks that demand fine-grained dual-arm coordination~\cite{openvla,rt1,palm-e,RDT-1B,robotwin}. Imitation learning (IL) has become the dominant paradigm for acquiring this skill~\cite{scar,bikc,D-CODA,ppi,peract2,stabilize,kstar,RDT-1B,deco}, and existing approaches fall into two main categories. The first is robot-centric IL, where models learn directly from demonstration trajectories: they either predict sparse robot keyposes for high-level planning~\cite{peract2,any-bimanual}, executed by motion planners~\cite{open-motion-planning,curobo}, or generate dense continuous action sequences for direct control~\cite{act,dp3}. Hybrid methods combine these paradigms~\cite{3d-diffuser-actor,3d_flowmatch_actor}, using keyposes to guide continuous action generation. More recently, object-centric IL methods~\cite{ppi,mba} enrich robot-centric IL by incorporating the manipulated object’s motion as auxiliary guidance.

Despite significant progress, existing approaches still face critical limitations. As shown in the upper part of Fig.~\ref{fig:teaser}, current paradigms struggle in the ``pick plate" task that requiring fine-grained dual-arm coordination (one arm tilting the plate while the other grasps it). Keypose-based methods enhance spatial awareness but, due to sparse supervision and poor control of intermediate states, often yield inaccurate trajectories and inter-arm collisions. Continuous action-based methods reproduce diverse motions, yet their reliance on dense supervision leads to overfitting, weakening perception and generalization. Hybrid methods combining keyposes and continuous actions seek a balance but fail to capture the object’s dynamic state, causing one arm to be unable to tilt the plate properly. Object-centric extensions enhance object awareness by leveraging movement information (e.g., object pointflow), but they overlook explicit robot–object interactions, which may lead to failures, such as a plate slipping before the other arm completes the grasp. Overall, these limitations make bimanual systems unstable in tasks that demand high-precision coordination.

In contrast, when performing bimanual operations, humans naturally sustain a continuous awareness of the \textbf{spatial triadic relationship}, that is, both the relation between the two hands and the relation between each hand and the manipulated object. Such triadic reasoning implicitly encodes inter-hand distances, hand–object relations, and their dynamic interactions, thereby ensuring stable and collision-free coordination.
Consistently, findings from UMI~\cite{umi} show that providing policies with the relative pose between grippers is crucial for achieving effective bimanual coordination. In the domain of dexterous manipulation, cutting-edge works such as $\mathcal{D(R,O)}$~\cite{dro} and RobustDexGrasp~\cite{RobustDexGrasp} have also demonstrated that explicitly modeling hand–object interactions can significantly improve manipulation performance. Taken together, we argue that constructing a reasonable spatial triadic interaction for bimanual system and incorporating it into bimanual policies can enhance manipulation stability.

Motivated by this insight, we introduce \textbf{RoTri-Diff}, a novel IL framework guided by the spatial \textbf{R}obot–\textbf{O}bject \textbf{Tr}iadic \textbf{I}nteraction (\textit{RoTri}). 
At its core, \textit{RoTri} models the triadic spatial configuration formed by the two end-effectors and the manipulated object. By uniformly encoding their relative 6D pose, it establishes a continuous triangular geometric constraint that provides structured guidance for feasible and stable bimanual coordination.
As shown in Fig.~\ref{fig:teaser}, RoTri-Diff is the first bimanual IL framework that integrates three key guidance signals: robot keyposes for long-horizon planning, object movement for capturing physical dynamics, and the core \textit{RoTri} representation for maintaining stable spatial relations. Architecturally, RoTri-Diff is a hierarchical diffusion model that can generate action sequences while maintaining consistency from global planning to fine-grained execution. The process includes three stages: (i) simultaneously predicting object pointflow and a continuous \textit{RoTri} segment; (ii) generating keypose actions based on the predicted object pointflow and the \textit{RoTri} at the keypose timesteps from (i); and (iii) integrating the three signals produced in (i) and (ii) to generate continuous action sequences.

We extensively evaluate RoTri-Diff in both simulation and the real world. On the RLBench2 bimanual benchmark~\cite{peract2}, it achieves an \textbf{10.2\% higher} success rate across 11 representative tasks compared to state-of-the-art baselines. Furthermore, in four challenging real-world bimanual manipulation tasks, RoTri-Diff demonstrates robust and stable performance.  

In summary, our contributions are threefold:  
\begin{itemize}
    \item We introduce the concept of \textbf{\textit{RoTri}} for bimanual manipulation, a triadic interaction representation that explicitly encodes spatial relations between the two arms and the object, enabling stable bimanual actions.
    \item We introduce \textbf{RoTri-Diff}, a hierarchical diffusion model that synergistically integrates robot keyposes, object dynamics, and \textit{RoTri} constraints to generate spatially and temporally consistent trajectories.  
    \item We provide extensive empirical validation, achieving state-of-the-art performance on 11 RLBench2 tasks and robust real-world execution in four bimanual manipulation scenarios.
\end{itemize}

\begin{table}[t!]
    \centering
    \resizebox{0.5\textwidth}{!}{
        \begin{tabular}{@{}l cccc@{}}
            \toprule
            & \multicolumn{3}{c}{\textbf{Imitation Learning Guides}} & \\
            \cmidrule(lr){2-4}
            \textbf{Method} 
            & \begin{tabular}[c]{@{}c@{}}Robot \\ Keyposes\end{tabular} 
            & \begin{tabular}[c]{@{}c@{}}Object \\ Movement\end{tabular} 
            & \begin{tabular}[c]{@{}c@{}}Robot--Object \\ Interaction\end{tabular} 
            & \textbf{Prediction Type} \\
            \midrule
            ACT \cite{act} & \xmark & \xmark & \xmark & Continuous Actions \\
            $\text{PerAct}^2$ \cite{peract2} & \cmark & \xmark & \xmark & Keyposes \\
            DP3 \cite{dp3} & \xmark & \xmark & \xmark & Continuous Actions \\
            AnyBimanual \cite{any-bimanual} & \cmark & \xmark & \xmark & Keyposes \\
            3D Diffuser Actor \cite{3d-diffuser-actor} & \cmark & \xmark & \xmark & Keyposes \& Continuous Actions\\
            PPI \cite{ppi} & \cmark & \cmark & \xmark & Keyposes \& Continuous Actions \\
            \textbf{RoTri-Diff (Ours)} & \cmark & \cmark & \cmark & Keyposes \& Continuous Actions \\
            \bottomrule
        \end{tabular}
    }
    \caption{Comparison of representative bimanual imitation learning methods.}
    \label{tab:relatedbimanual}
    \vspace{-5mm}
\end{table}

\section{Related Work}

\subsection{Imitation Learning for Bimanual Manipulation}

Current imitation learning (IL) approaches for bimanual manipulation can be broadly categorized into two paradigms~\cite{act,any-bimanual,stabilize,peract2}. 
The first is robot-centric, which directly models and predicts robot states. It includes: (i) keypose-based methods~\cite{peract2,any-bimanual} that predict discrete waypoints and rely on motion planners for execution; (ii) continuous action-based methods~\cite{act,RDT-1B,dp3} that generate dense control sequences using Transformers or diffusion models; and (iii) hybrid methods~\cite{3d-diffuser-actor,3d_flowmatch_actor} that combine high-level keypose prediction with low-level trajectory generation.
The second is object-centric, which uses object motion as the primary guidance signal for action generation~\cite{ppi,mba}. For instance, PPI~\cite{ppi} leverages 3D point-level object flow to enhance spatial grounding and manipulation precision.
Despite their differences, both paradigms fail to explicitly model the dynamic spatial relationship between the two arms and the object, limiting performance in contact-rich tasks requiring tight coordination and collision avoidance. To address this gap, we introduce the \textit{RoTri} representation as a unified guidance signal. As shown in Table~\ref{tab:relatedbimanual}, our method integrates robot-object interactions, robot keyposes, and object motion within a hierarchical framework, enabling robust prediction of both end-effector keyposes and continuous actions in complex bimanual scenarios.

\begin{figure*}[t!]
    \centering
    \includegraphics[width=\textwidth]{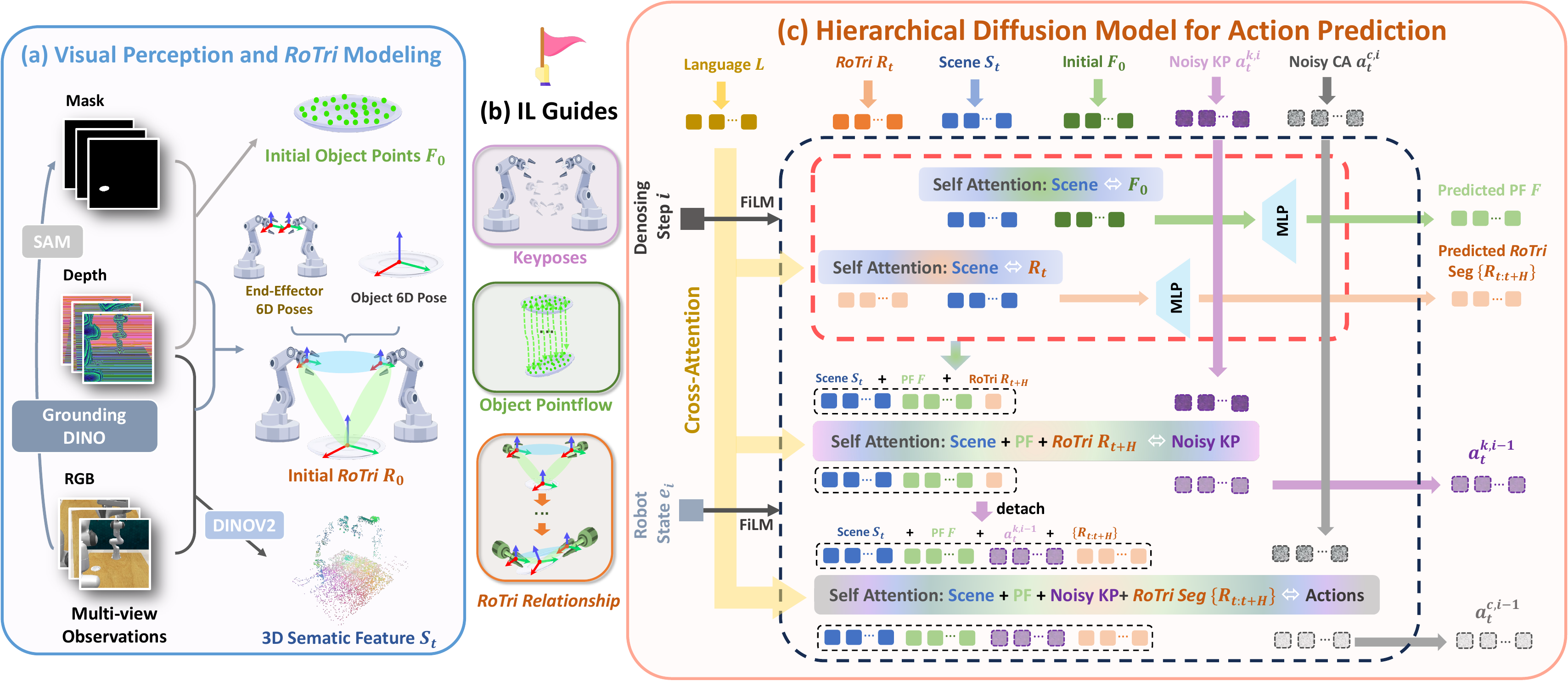}
    \caption{\textbf{Overview of RoTri-Diff.} 
     (a) \textbf{Visual Perception and \textit{RoTri} Modeling:} Extracting the initial object point cloud $F_0$, 3D semantic features $S_t$, and the initial \textit{RoTri} representation $R_0$ from multi-view observations. 
    (b) \textbf{Imitation Learning Guidance Signals:} Three complementary signals used for supervision: Keyposes, Object Pointflow, and the \textit{RoTri} Relationship. 
    (c) \textbf{Hierarchical Diffusion Model:} The model concurrently predicts object pointflow and autoregressively predicts a future \textit{RoTri} segment. These predictions then serve as dynamic conditions to guide the denoising and generation of keyposes and continuous actions within a synergistic attention module.
    }
    \label{fig:method}
    \vspace{-5mm}
\end{figure*}

\subsection{Diffusion Models for Manipulation}
Diffusion models, such as Denoising Diffusion Probabilistic Models (DDPM)~\cite{ddpm}, have achieved remarkable success in generative domains including image~\cite{sdxl,rombach2022high} and video synthesis~\cite{prolificdreamer,scalable}. Inspired by these advances, recent work has applied diffusion models to robotic action generation via imitation learning. These methods typically learn policies that iteratively denoise action trajectories, conditioned on perceptual inputs from 2D images~\cite{diffusion,dp,hdp,render-and-diffuser,sdp} or 3D point clouds~\cite{3d-diffuser-actor,dp3,gravmad,gdp3}.  
In the context of bimanual manipulation, diffusion-based approaches have further explored unified action spaces~\cite{RDT-1B}, constrained image-conditioned policies~\cite{D-CODA}, kinematics-aware action generation~\cite{kstar}, and object-flow guidance~\cite{ppi}. While these advances mark important progress, they still inherit the key limitation outlined above: the lack of explicit modeling of the spatial triadic interactions between the arms and the object.  
In contrast, our proposed RoTri-Diff is the first diffusion-based framework to explicitly incorporate this relational geometry. By conditioning the denoising process on the \textit{RoTri} representation, our hierarchical model significantly improves the stability and coherence of generated bimanual actions.  

\section{Method\label{sec:method}}
In this section, we detail the overall architecture of \textbf{RoTri-Diff}. We start with a brief problem formulation (Sec.~\ref{method:formul}), then describe visual perception and Robot-Object Triadic Interaction (\textit{RoTri}) modeling (Sec.~\ref{method:perception_and_model}), including 3D semantic feature construction, initial object point cloud acquisition, and building the unique \textit{RoTri} representation from the 6D poses of the two arm end-effectors and the object. Next, we cover key imitation learning (IL) guidance signals: keyposes, object pointflow, and the \textit{RoTri} relationship (Sec.~\ref{method:guides}). We then present the action prediction module (Sec.~\ref{method:hdm}), a hierarchical, diffusion-based Transformer with synergistic attention for precise action generation. Finally, we detail the training and inference procedures (Sec.~\ref{method:details}). Fig.~\ref{fig:method} provides an overview of our proposed framework.

\subsection{Problem Formulation}\label{method:formul}
Given a set of bimanual manipulation demonstrations $\mathcal{D} = \{(o_t, a_t)_{t=0}^{T_i}\}_{i=0}^N$, our goal is to train a RoTri-Diff network $\epsilon_{\theta}$ that maps visual observations $o$ to actions $a$. At timestep $t$, our model takes as input the language instruction and RGBD images from $K$ cameras, and outputs a sequence of $h^c$ continuous actions $a_t^c = \{a_{t:t+h^c-1}\}$, where each action $a_t$ represents the target gripper poses and openness for both the left and right manipulators.
The model incorporates three key imitation learning (IL) guidance signals: Keyposes, Object Pointflow, and the Robot-Object Triadic Interaction (\textit{RoTri}) representation. A keypose timestep $t^k$ is defined as a turning point in the trajectory with significant changes in the grippers' openness and the arms' joint states~\cite{qattention,peract2,3d-diffuser-actor,gravmad}. 
The first signal, Keyposes, specifies the target end-effector poses for the subsequent $h^k$ keyposes, $a_t^k = \{a_{t^k_i}\}_{i=1}^{h^k}$. The second, Object Pointflow, defines the positions of $N_q$ object point cloud points at the next $h^k$ keyposes, $F \in \mathbb{R}^{h^k \times N_q \times 3}$. Finally, the third signal, the \textit{RoTri} representation $R_t \in \mathbb{R}^{21}$, precisely quantifies the 6D relative poses of the two arms and the object.

\subsection{Visual Perception and \textit{RoTri} Modeling}\label{method:perception_and_model}
\textbf{Visual Perception.}
As is seen in Fig.~\ref{fig:method}(a), to obtain 3D semantic features that integrate both semantic and geometric information, we adopt the 3D scene representation method from PPI~\cite{ppi}. We first preprocess the raw point clouds through cropping and downsampling. We then use the DINOv2 model~\cite{dinov2} to extract pixel-wise semantic features from multi-view 2D images. These features are fused via a weighted sum, with weights determined by the point's distance from the projected surface. To mitigate computational burden, we use a PointNet++ dense encoder~\cite{pointnet} to downsample the scene points into a compact representation $S_t \in \mathbb{R}^{N_s \times (3+D)}$. This compressed representation retains key geometric and semantic details while enhancing local point relationships through the set abstraction of PointNet++.

To enable the model to focus on capturing overall object motion rather than global absolute coordinates, thereby ensuring robust and generalized pointflow estimation even when the object's position is out of distribution, we use the same initial object point cloud sampling strategy as in PPI~\cite{ppi}, which approximates the conditional distribution $p(F|F_0)$. Here, $F_0$ represents the initial point cloud sampled from the manipulated object in the first frame. To acquire $F_0$, we use the Grounding DINO~\cite{groundingdino} and SAM~\cite{sam} models to generate an object mask from a language prompt and an image. We then randomly sample $N_q=200$ pixels from the mask and obtain their 3D coordinates $F_0 \in \mathbb{R}^{N_q \times 3}$. Consistent with the setup in PPI~\cite{ppi}, this sampling operation is performed only once per episode.

\textbf{\textit{RoTri} Modeling.}
The core of our method is to capture the dynamic triadic interaction formed by the two arms and the manipulated objects. We define the Robot-Object Triadic Interaction (\textit{RoTri}) vector as a compact representation of this relationship. To obtain the initial \textit{RoTri} vector, we first acquire the 6D absolute poses of the two arm end-effectors and the object at the initial timestep. We then compute the relative poses between each pair and concatenate them into a single vector:
$R_0 = [p^0_{\text{left} \to \text{right}},\; p^0_{\text{left} \to \text{obj}},\; p^0_{\text{right} \to \text{obj}}] \in \mathbb{R}^{21},$
where each 7-dimensional component consists of a 3D position and a 4D quaternion ($x, y, z, w$). For training supervision, this representation is computed at every step of the expert trajectory, yielding the ground-truth sequence $\{R_t\}_{t=0}^T$.
For tasks with multiple manipulated objects, we extend \textit{RoTri} modeling to handle variable interactions while maintaining a fixed-size representation. We construct a set of \textit{RoTri} vectors, each describing the triadic interaction with one object, and aggregate them using a permutation-invariant Transformer encoder. The encoder outputs a high-dimensional embedding summarizing the most relevant interactions, which is then projected back into a fixed 21-dimensional vector. This allows the model to dynamically focus on the relevant objects while remaining scalable and robust to varying object counts.

\subsection{IL Guidance Signals}\label{method:guides}
\textbf{Keyposes \& Object Pointflow.} 
Following PPI~\cite{ppi}, we generate two key IL guidance signals. First, we employ the heuristic algorithm~\cite{qattention} to extract keyposes $a_{t_i^k}$ from expert trajectories, which are defined as points where joint velocity approaches zero or gripper open/close events occur, and these serve as targets to guide continuous action generation. Second, to address the occlusion issue caused by object motion, we leverage the object's 6D pose (obtained from simulation~\cite{peract2} or the real world~\cite{bundlesdf,foundationpose}) to transform initial object point cloud $F_0$ to their future keypose positions $F_{t_i^k} \in \mathbb{R}^{N \times 3}$. These transformed points serve as the ground truth for pointflow supervision and eliminate the need for real-time pose estimation during inference.

\textbf{\textit{RoTri} Relationship.} 
Instead of directly predicting the complete \textit{RoTri} trajectory, we adopt a more robust approach by learning its dynamic evolution through approximating the conditional distribution $p(R|R_0)$. Specifically, the model predicts in an incremental fashion: at each timestep, it predicts the change in the \textit{RoTri} representation, $\Delta R_t$. The subsequent \textit{RoTri} representation $R_t$ is then obtained by accumulating this predicted change onto the previous timestep's representation, i.e., $R_t = R_{t-1} + \Delta R_t$.
This predictive paradigm shifts the learning focus from inferring complex absolute poses to capturing relative interaction dynamics.

\subsection{Hierarchical Diffusion Model for Action Prediction}\label{method:hdm}

\textbf{Observation Encoder.}
The observation encoder of RoTri-Diff processes and encodes multi-modal inputs, including the 3D semantic features, language instruction, robot states, initial object points, and the initial \textit{RoTri} representation into a unified latent space. The language instruction is processed by a CLIP-based text encoder, while the robot states (which include pose and gripper state for both arms), initial object points, and the initial \textit{RoTri} representation are each independently projected into the latent space via separate three-layer MLPs. These encoded features are then integrated to form the final observational conditioning for the model.

\textbf{Action Prediction.}
As shown in Fig.~\ref{fig:method}(c), at timestep $t$ and denoising step $i$, RoTri-Diff integrates \textit{RoTri} tokens $\textit{R}_t$, scene tokens $S_t$, language tokens $L$, initial object points tokens $F_0$, and noised action tokens $a_t^{k,i}$ and $a_t^{c,i}$. Each token comprises a latent embedding and a 3D position. Following 3D Diffuser Actor~\cite{3d-diffuser-actor} and PPI~\cite{ppi}, we apply Relative Self-Attention with Rotary Positional Embedding to effectively leverage the relative 3D spatial information between tokens. 
The robot's proprioception $e_t$ and denoising timesteps $i$ affect the attention through Feature-wise Linear Modulation (FiLM)~\cite{film}.
All tokens first attend to the language tokens $L$ via Parallel Attention. Subsequently, the model initiates two parallel processes: \textbf{1) Object pointflow (PF) prediction:} In a one-shot process, initial object points tokens $F_0$ attend to scene tokens $S_t$. An MLP then predicts a 12D representation (xyz coordinates for subsequent keyposes) for each object points. \textbf{2) Auto-regressive \textit{RoTri} prediction:} This process follows a hierarchical auto-regressive prediction mechanism, iterating on a keypose-by-keypose basis. The model auto-regressively predicts the \textit{RoTri} segment 
$\{R_{t:t+H}\}$ for the next $H$ timesteps. Each prediction is conditioned on the previous keypose step's \textit{RoTri} state (initially $R_0$) attending to scene tokens and immediately serves as a dynamic condition for action generation.

Rather than a single-shot prediction, the model proceeds step-wise across keypose segments, predicting each segment’s \textit{RoTri} trajectory conditioned on the previous segment’s final \textit{RoTri} (or $R_0$ for the first segment).
The predicted \textit{RoTri} segment conditions keypose action generation. Noised keypose tokens $a_t^{k,i}$ attend to the object PF, scene, and the specific \textit{RoTri} token at the keypose timestep ($\textit{R}_{t+H}$) via Relative Self-Attention. An MLP maps the output to the action dimension, supervised by keypose noise $\epsilon_t^{k,i}$. Then, noised continuous action tokens $a_t^{c,i}$ attend to all prior features, including the new keypose and full \textit{RoTri} segment. Another MLP predicts continuous noise $\epsilon_t^{c,i}$ for supervision. The final state of the predicted \textit{RoTri} segment $\textit{R}_{t+H}$ feeds into the next iteration, completing the auto-regressive chain.

\subsection{Implementation Details}\label{method:details}
\textbf{Training.} \quad For the RoTri-Diff network $\epsilon_{\theta}$, all losses are computed via an $L_1$ norm. Given the conditioning inputs (3D features $S_t$, language $L$, initial points $F_0$, initial \textit{RoTri} $R_0$, and proprioception $e_t$), the loss functions are:
$$
\begin{aligned}
    \mathcal{L}_c &= ||\epsilon_{\theta}(S_t, L, F_0, R_0, e_t, a_t^{c,i}, i) - \epsilon_c||_1 \\
    \mathcal{L}_k &= ||\epsilon_{\theta}(S_t, L, F_0, R_0, e_t, a_t^{k,i}, i) - \epsilon_k||_1 \\
    \mathcal{L}_{pf} &= ||\epsilon_{\theta, \text{pf}}(S_t, L, F_0, e_t, i) - F||_1 \\
    \mathcal{L}_{\text{rotri}} &= ||\epsilon_{\theta, \text{rotri}}(S_t, L, R_0, e_t, i) - \Delta R_t||_1
\end{aligned}
$$
Here, $a_t^{c,i}$ and $a_t^{k,i}$ are noised actions, while $\epsilon_c$ and $\epsilon_k$ are the ground truth noise. $\epsilon_{\theta, \text{pf}}$ and $\epsilon_{\theta, \text{rotri}}$ denote the prediction heads for the object PF $F$ and the change in RoTri representation $\Delta R_t$.
The total training loss is: 
\[ \mathcal{L}_{\theta} = w_c \mathcal{L}_{c} + w_k \mathcal{L}_{k} + w_{pf} \mathcal{L}_{pf} + w_{\text{rotri}} \mathcal{L}_{\text{rotri}}, \]
with weights $w_c, w_k = 0.05$ and $w_{pf}, w_{\text{rotri}} = 1.0$. 
In all experiments, we use a DDPM noise schedule with 1000 training timesteps. For tasks from the RLBench benchmark, we train for 500 epochs; for real-world tasks, we train for 1000 epochs. We use a batch size of 64, an AdamW optimizer with a learning rate of 1e-4, and a cosine decay learning rate scheduler, and conduct model training on eight A5000 GPUs. We select the checkpoint with the lowest average validation loss for evaluation.

\vspace{\baselineskip}
\textbf{Inference.} \quad At the start of an episode ($t=0$), we sample $N_q=200$ object points $F_0$ and compute the initial \textit{RoTri} representation $R_0$, which serve as global conditions. At each key decision point $t$, the policy generates actions by starting with random noise $\hat{a}_t^N \sim \mathcal{N}(0, \mathbf{I})$ and performing an iterative hierarchical diffusion process. In each denoising step, the model predicts the future PF $F$ and \textit{RoTri} segment $\{R_{t:t+H}\}$ as dynamic guidance. After a set number of steps (1000 for DDPM in sim, 20 for DDIM in real), the process yields a clean sequence of 50 continuous actions. The entire generated action sequence is then sent to the robot controller for complete execution until the next keypose is reached, at which point the planning cycle repeats.

\section{Experiments}
\begin{figure}
    \centering
    \includegraphics[width=\linewidth]{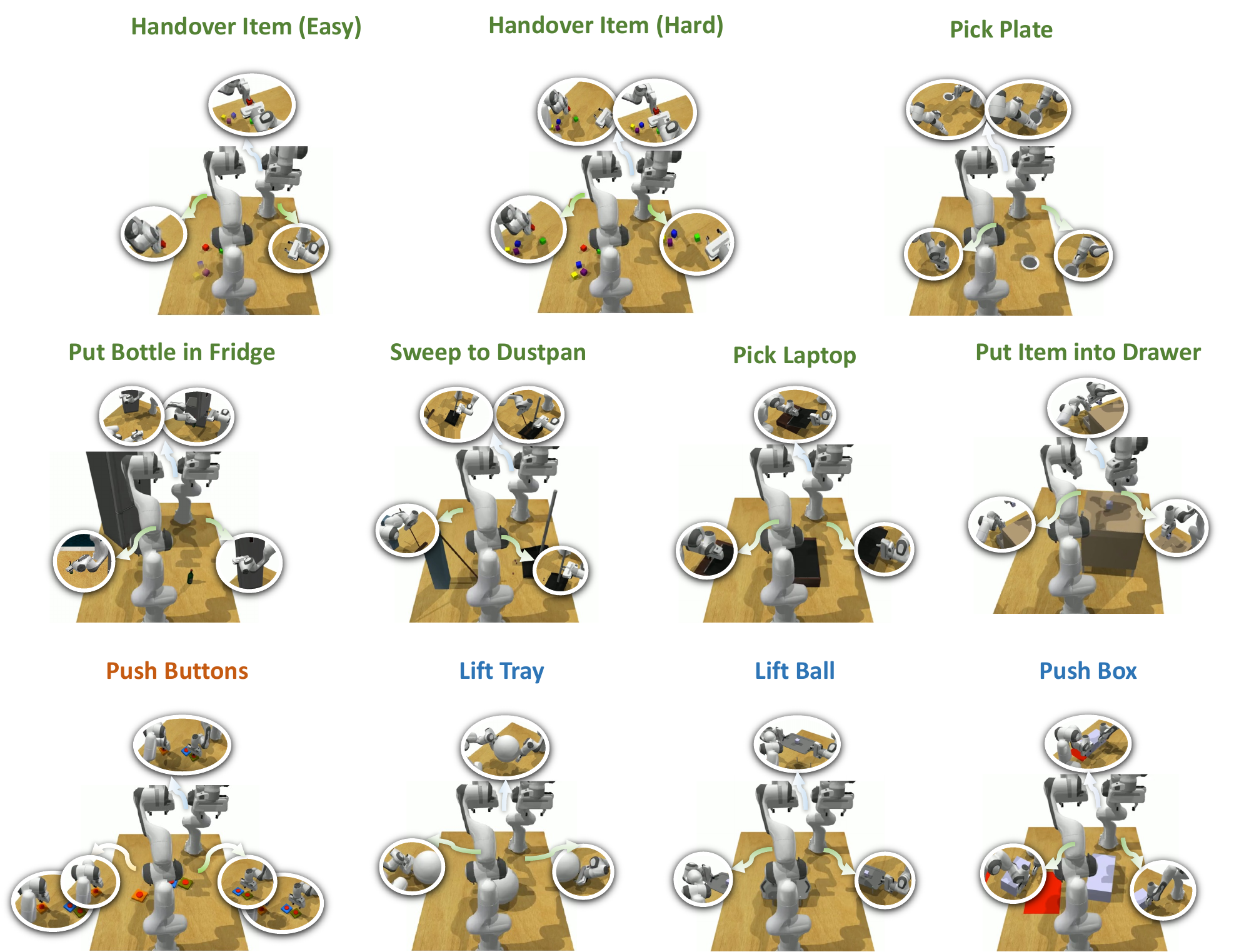}
    \caption{Visualization of the 11 RLBench2 tasks.}
    \vspace{-5mm}
    \label{fig:sim_exp_setup}
\end{figure}

\begin{table*}[t]
    \centering
    \begin{adjustbox}{width=\textwidth}
    \Large
    \begin{tabular}{@{}l*{12}{c}@{}}
    \toprule
    & Avg. Success (\%)
    & \makecell{push \\ buttons} 
    & \makecell{lift \\ ball} 
    & \makecell{lift \\ tray} 
    & \makecell{push \\ box} 
    & \makecell{handover \\ item (easy)} 
    & \makecell{handover \\ item (hard)} 
    & \makecell{pick \\ laptop} 
    & \makecell{put item \\ into drawer} 
    & \makecell{sweep \\ to dustpan} 
    & \makecell{pick \\ plate} 
    & \makecell{bottle \\ in fridge} \\
    \midrule
    ACT~\cite{act} & 5.9
    & 4.0 & 38.3 & 1.3 & 67.0 & 0.0 & 0.0 & 0.0 & 1.7 & 0.0 & 0.0 & 0.0\\
    PerAct$^2$~\cite{peract2} & 16.8
    & 47.0 & 50.0 & 60.0 & 62.0 & 19.7 & 11.0 & 36.7 & 49.7 & 2.0 & 4.0 & 3.0 \\
    DP3~\cite{dp3} & -
    & - & 27.0 & 6.3 & 39.3 & 4.7 & - & 6.0 & 0.0 & 98.7 & - & - \\
    AnyBimanual~\cite{any-bimanual} &  39.9
    & 73.0 & 36.0 & 67.0 & 46.0 & 44.0 & 15.0 & 7.0 & 50.0 & 67.0 & 8.0 & 26.0 \\
    3DDA~\cite{3d-diffuser-actor} & -
    & - & 87.3 & 76.0 & 54.7 & 44.7 & - & 40.7 & 52.7 & 96.7 & - & - \\
    PPI~\cite{ppi} &  70.7
    & 92.0 & 89.3 & 92.0 & \textbf{96.7} & 62.7 & 37.3 & 46.3 & 79.7 & \textbf{98.7} & 0.0 & 82.6 \\
    \textbf{RoTri-Diff (ours)} & \cellcolor{ImportantColor} \textbf{80.9} \textcolor{red}{(10.2\%$\uparrow$)} 
    & \cellcolor{ImportantColor} \textbf{97.0$_{\pm 1.73}$} & \cellcolor{ImportantColor} \textbf{95.7$_{\pm 1.15}$} & \cellcolor{ImportantColor} \textbf{94.3$_{\pm 1.15}$} & \cellcolor{ImportantColor} 95.0$_{\pm 1.73}$ & \cellcolor{ImportantColor} \textbf{73.3$_{\pm 2.89}$} & \cellcolor{ImportantColor} \textbf{52.3$_{\pm 2.52}$}  & \cellcolor{ImportantColor} \textbf{66.0$_{\pm 3.46}$} & \cellcolor{ImportantColor} \textbf{87.0$_{\pm 1.73}$} & \cellcolor{ImportantColor} 96.67$_{\pm 2.89}$ & \cellcolor{ImportantColor} \textbf{40.7$_{\pm 1.15}$} & \cellcolor{ImportantColor} \textbf{92.0$_{\pm 1.73}$} \\
    \bottomrule
    \end{tabular}
    \end{adjustbox}
    \caption{\textbf{Performance on 11 RLBench2 tasks.}  
    AnyBimanual is a multi-task policy evaluated with a single checkpoint across all tasks, whereas the other methods are single-task policies that report results from the best checkpoint for each task~\cite{ppi,3d_flowmatch_actor}. RoTri-Diff is evaluated with three seeds per task, reporting the mean and standard deviation.}
    \vspace{-3mm}
    \label{tab:main}
\end{table*}

In this section, we perform a series of experiments aimed at addressing the following questions:
\begin{itemize}[leftmargin=5pt]
\item[] \textbf{Q1:} How well does RoTri-Diff perform on complex bimanual manipulation tasks in simulation, and can it outperform state-of-the-art methods?
\item[] \textbf{Q2:} What critical role does the \textit{RoTri} representation play? Among keypose guidance, continuous action prediction, and the synergy between the two, which application of \textit{RoTri} is most critical for improving model performance?
\item[] \textbf{Q3:} Can RoTri-Diff maintain stable performance on real-world bimanual tasks that require precise spatial coordination and strict action constraints?
\end{itemize}

\subsection{Simulation Results}

\textbf{Simulation Setup.}
To answer \textbf{Q1}, we perform extensive simulation experiments in RLBench2~\cite{peract2}, a bimanual manipulation benchmark built on CoppeliaSim featuring diverse tasks with varying coordination demands.
As shown in Fig.~\ref{fig:sim_exp_setup}, we select eleven challenging tasks, grouped into three bimanual coordination types~\cite{peract2}:
1) \textbf{Symmetric Coordination}: Both arms move simultaneously on separate objects. The challenge lies in parallel operation within a shared workspace, requiring continuous perception of spatial relations to plan collision-free trajectories. Task: \textit{Push Buttons}.
2) \textbf{Synchronous Coordination}: Both arms jointly manipulate a single object. The difficulty is precise coordination of relative poses to prevent drops or collisions. Tasks: \textit{Lift Tray}, \textit{Lift Ball}, \textit{Push Box}.
3) \textbf{Asynchronous Coordination}: Arms perform dependent sequential actions on the same object. The challenge is capturing accurate spatiotemporal interaction to ensure reliable execution while avoiding drops or collisions. Tasks: \textit{Put Button in Fridge}, \textit{Sweep to Dustpan}, \textit{Pick Laptop}, \textit{Put Item into Drawer}, \textit{Handover Item (Easy)}, \textit{Handover Item (Hard)}, \textit{Pick Plate}.

\textbf{Baselines.}
We compare RoTri-Diff with a diverse set of baselines to highlight its advantages across different levels. 
ACT~\cite{act} and DP3~\cite{dp3} directly predict continuous action sequences without leveraging high-level guidance. 
PerAct$^2$~\cite{peract2} and AnyBimanual~\cite{any-bimanual} employ a Perceiver-based architecture~\cite{perceiverio} to predict discrete keyposes. 
3D Diffuser Actor (3DDA)~\cite{3d-diffuser-actor} further integrates diffusion to predict both keyposes and full action trajectories. 
PPI~\cite{ppi} combine keypose and object-movement guidance, achieving strong performance on tasks with dynamic objects. 
DP3 and 3DDA are evaluated on only a subset of the 11 tasks.

\textbf{Metric.}
Our method is evaluated with 100 rollouts per task under randomized initial states. We report the success rate (\%) for each task as well as the average across all tasks, where results are averaged over three independent checkpoints and presented with mean and standard deviation.

\textbf{Overall Performance.}
As shown in Table~\ref{tab:main}, our RoTri-Diff model achieves an average success rate of \textbf{80.9\%} across 11 tasks. This result not only significantly outperforms all baselines but also represents a \textbf{10.2\% improvement} over the previous state-of-the-art.
Methods that rely solely on continuous action prediction, such as ACT and DP3, perform poorly on tasks requiring fine-grained control; for instance, DP3 achieves only 0.0\% on \textit{put item into drawer}. Similarly, keypose-based methods like PerAct$^2$ and AnyBimanual struggle with challenging coordination tasks. On \textit{handover item (hard)}, for example, AnyBimanual achieves only 15.0\%, whereas our RoTri-Diff reaches 52.3\%.
Although 3DDA combines keyposes with continuous actions, it lacks explicit modeling of object dynamics. As a result, it underperforms significantly in asynchronous coordination tasks: on \textit{handover item (easy)}, it achieves only 44.7\%, compared to RoTri-Diff’s 73.3\%. Similarly, on \textit{put item into drawer}, RoTri-Diff achieves 87.0\%, far exceeding 3DDA’s 52.7\%.

Against the state-of-the-art hybrid baseline PPI, our RoTri-Diff demonstrates clearer advantages. While PPI is the first to introduce object movement guidance, it completely fails on the highly constrained \textit{pick plate} task (0.0\%), whereas RoTri-Diff achieves 40.7\%. On \textit{handover item (easy)} and \textit{put item into drawer}, RoTri-Diff also surpasses PPI by large margins, reaching 73.3\% vs. 62.7\% and 87.0\% vs. 79.7\%, respectively.
In summary, by explicitly modeling the \textit{RoTri} relationship, our method enables the system to directly perceive the spatial relation among the two arms and the manipulated object, thereby enhancing its performance in tasks that require precise coordination and spatial interaction.

\begin{table}[t]
    \centering
    \begin{adjustbox}{width=0.85\linewidth} 
    \begin{tabular}{@{}l*{4}{c}@{}}
    \toprule
    Method & \makecell{push \\ buttons} & \makecell{lift \\ tray} &  \makecell{pick \\ plate} & \makecell{bottle \\ in fridge} \\
    \midrule
    \textit{RoTri} (Keypose Only) & 93.7 & 92.6 & 29.3 & 90.3 \\
    \textit{RoTri} (Continuous Only) & 95.3 & 88.7 & 21.3 & 86.0  \\ 
    \midrule
    \textbf{RoTri-Diff} & \textbf{97.0} & \textbf{94.3} & \textbf{40.7} & \textbf{92.0}\\
    \bottomrule
    \end{tabular}
    \end{adjustbox}
    \caption{\textbf{Ablation Study.} Comparison of two architectural variants with RoTri-Diff, confirming the necessity of combining keypose and continuous \textit{RoTri} guidance in a full hierarchical design.}
    \vspace{-2mm}
    \label{tab:ablation}
\end{table}

\begin{figure}
    \centering
    \includegraphics[width=\linewidth]{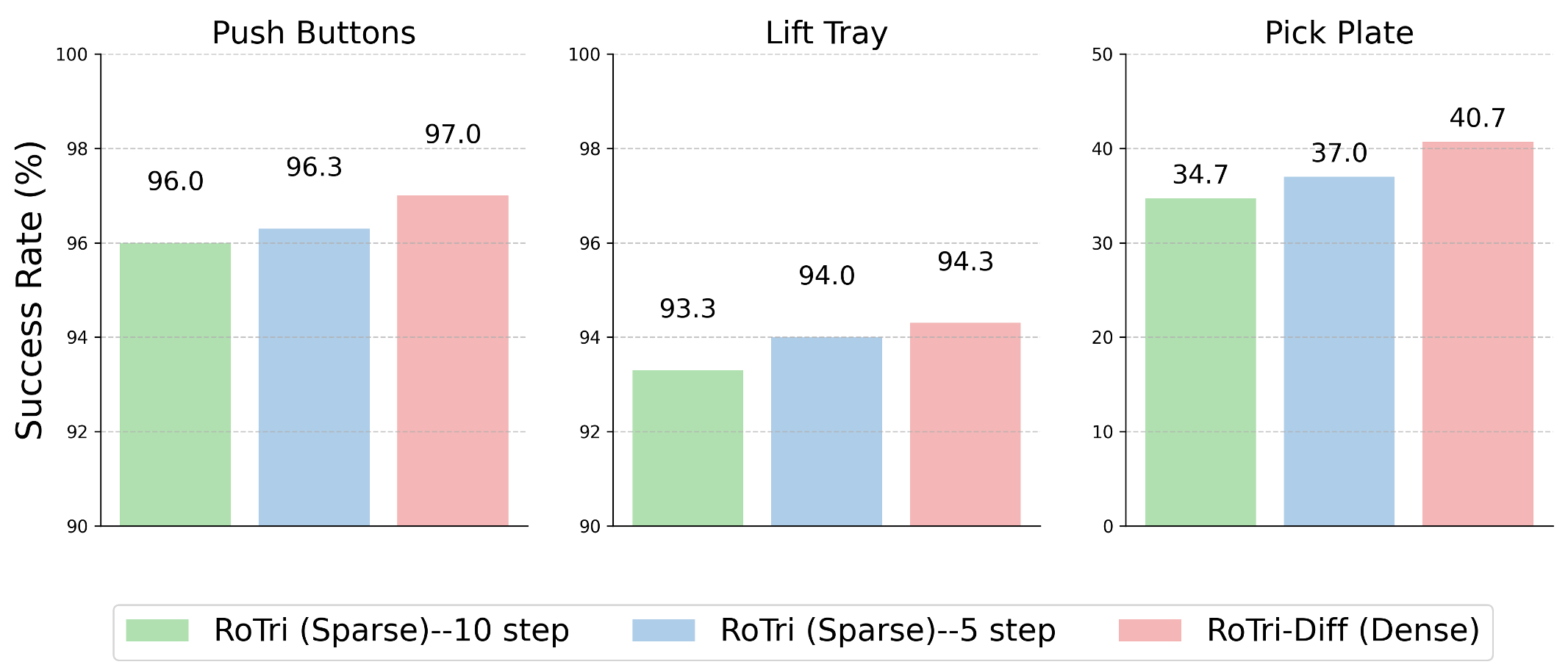}
    \caption{\textbf{Ablation Study.} Comparing Dense with Sparse (5/10 step) variants shows that higher guidance density consistently yields superior task success.}
    \vspace{-5mm}
    \label{fig:ablation2}
\end{figure}

\textbf{Ablation Study.} 
To address \textbf{Q2}, we conduct three ablation studies to examine the role of \textit{RoTri} representations and their synergy with the hierarchical prediction architecture. We design three variants:
\textbf{\textit{RoTri} (Keypose Only):}
This variant uses \textit{RoTri} only at the keypose timestep to guide high-level keypose prediction, while continuous actions are generated from predicted keyposes, scene features, and point cloud flow. It tests whether \textit{RoTri} mainly benefits high-level planning or also continuous execution.
\textbf{\textit{RoTri} (Continuous Only):}
This variant removes the keypose prediction module and directly uses the full \textit{RoTri} relationship to generate continuous action sequences, examining whether \textit{RoTri} alone can replace the hierarchical structure.
\textbf{\textit{RoTri} (Sparse):}
This variant studies the effect of temporal density by providing \textit{RoTri} guidance at fixed intervals (every 5 or 10 timesteps) between keyposes, evaluating how dense step-wise guidance affects performance.

As shown in Table~\ref{tab:ablation}, both representation and hierarchy are essential. In the Keypose Only variant, performance on \textit{Pick Plate} drops to 29.3\% (vs. 40.7\% for RoTri-Diff), showing that continuous constraints are needed for smooth, fine-grained actions. Conversely, the Continuous Only variant reaches 88.7\% on \textit{Lift Tray} but only 21.3\% on \textit{Pick Plate}, suggesting keyposes provide a crucial predictive anchor for complex tasks.
Fig.~\ref{fig:ablation2} further highlights the importance of temporal guidance density. RoTri-Diff consistently outperforms sparse variants: for example, on \textit{Push Buttons}, success drops from 97.0\% (dense) to 96.3\% (5-step) and 96.0\% (10-step), with larger gaps in precision-heavy tasks such as \textit{Pick Plate} (40.7\% vs. 37.0\%). This shows that dense per-timestep \textit{RoTri} guidance prevents drift and error accumulation.
Overall, the ablation confirms that combining explicit \textit{RoTri} representation, hierarchical prediction, and dense temporal guidance is key to robust bimanual coordination.

\begin{figure}[h]
    \centering
    \includegraphics[width=0.8\linewidth]{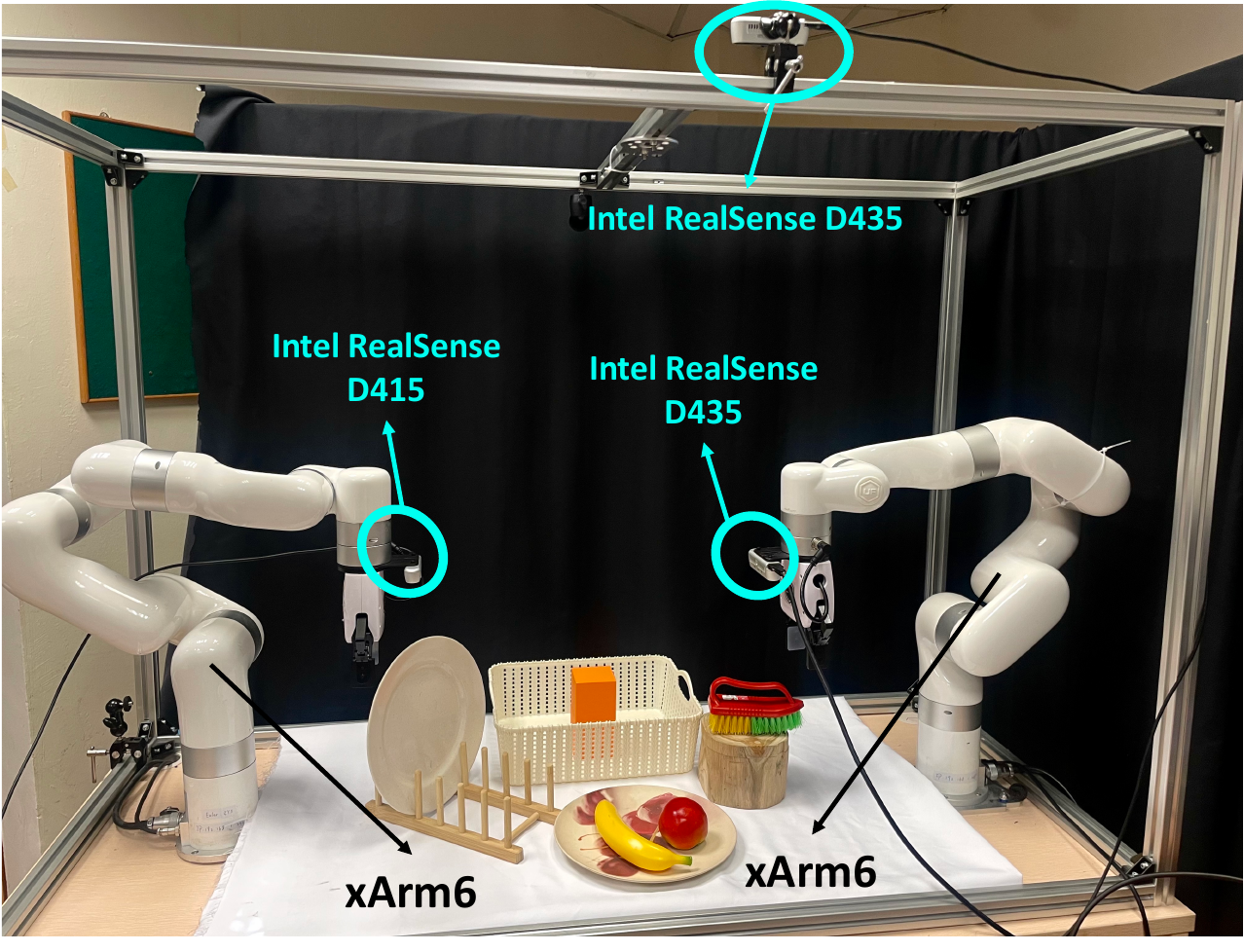}
    \vspace{-2mm}
    \caption{Real-world experimental setup.}
    \vspace{-7mm}
    \label{fig:real_exp_setup}
\end{figure}

\subsection{Real-world Results}
\begin{figure}[h]
    \centering
    \includegraphics[width=\linewidth]{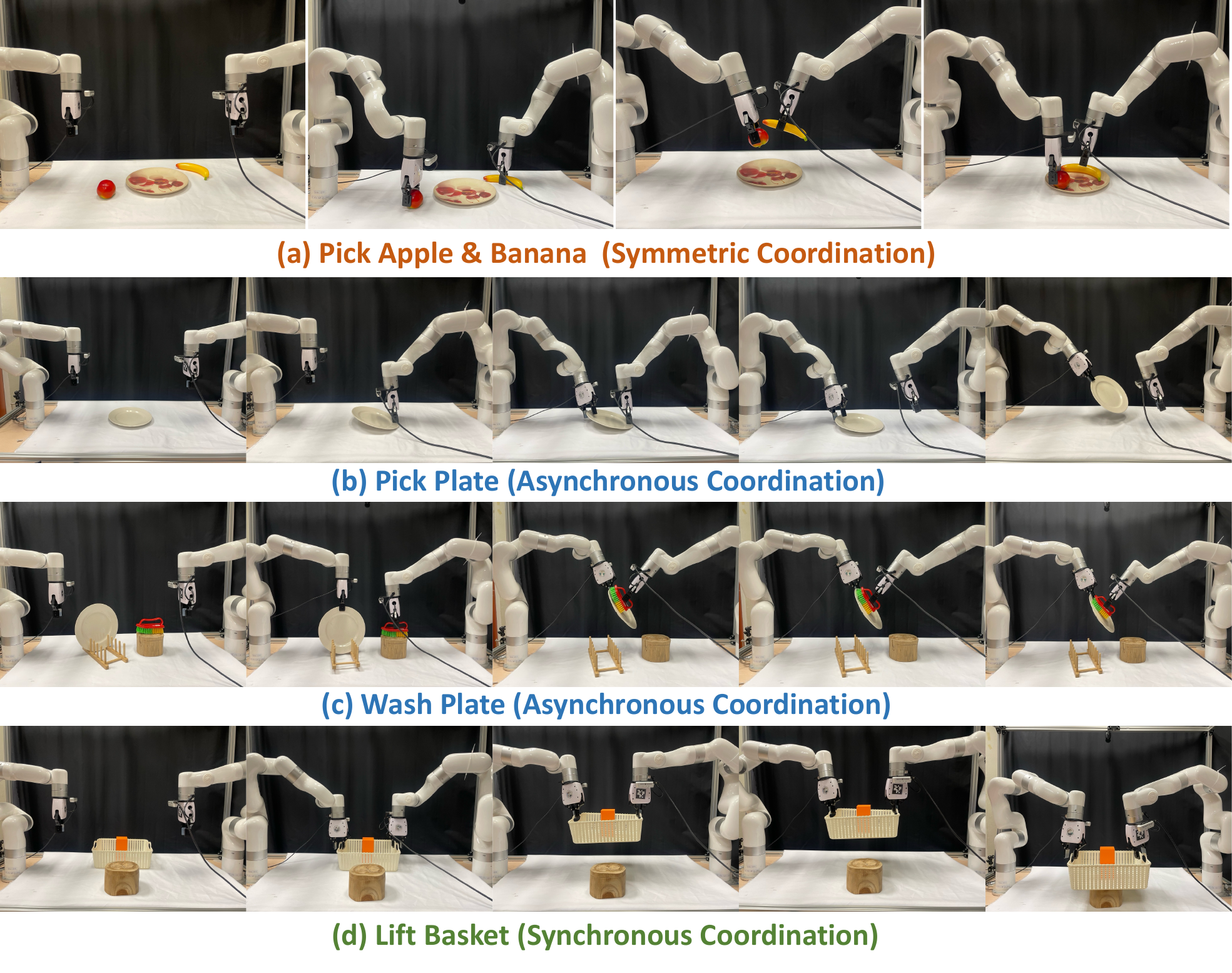}
    \vspace{-5mm}
    \caption{Task process visualizations in 4 real-world tasks.}
    \vspace{-5mm}
    \label{fig:real_4_tasks}
\end{figure}

\textbf{Real-world Setup.}
To answer \textbf{Q3}, we evaluate the proposed method on two xArm6 robots, with the setup shown in Fig.~\ref{fig:real_exp_setup}. Each scene includes two Eye-on-Hand cameras and one Eye-on-Base RealSense camera. Four real-world tasks are designed, covering the three coordination types:
1) \textbf{Pick Tomato \& Banana} (Symmetric Coordination): Both arms simultaneously grasp a tomato and a banana from the table and place them into a fruit bowl.  
2) \textbf{Pick Plate} (Asynchronous Coordination): One arm tilts the plate by pressing its edge, while the other grasps the lifted side.  
3) \textbf{Wash Plate} (Asynchronous Coordination): One arm retrieves a plate from the shelf and positions it, while the other grasps a brush and wipes the plate.  
4) \textbf{Lift Basket} (Synchronous Coordination): Both arms collaboratively lift a basket with stacked blocks and place it onto a wooden post while keeping the blocks upright.
The task processes are illustrated in Fig.~\ref{fig:real_4_tasks}. Demonstrations are collected via the Meta Quest 2 teleoperation platform: 50 for ``Lift Basket'' and ``Pick Place'', and 20 for the other tasks.

\textbf{Metric.} We report the total number of successful outcomes of RoTri-Diff across four real-world manipulation tasks, summing the successful trials (1 if the task is completed, 0 otherwise) over 5 trials per task. This total reflects overall performance across all trials.

\begin{table}[h]
    \centering
    \begin{adjustbox}{width=0.8\linewidth}
    \begin{tabular}{@{}l*{4}{c}@{}}
    \toprule
    \textbf{Method} & \makecell{Pick \\ Tomato \& Banana} & \makecell{Pick \\ Plate} & \makecell{Wash \\ Plate} & \makecell{Lift \\ Basket} \\
    \midrule
    \textbf{RoTri-Diff} & 5/5 & 3/5 & 4/5 & 4/5 \\
    \bottomrule
    \end{tabular}
    \end{adjustbox}
    \caption{Real-world experiment results.}
    \label{tab:real}
    \vspace{-3mm}
\end{table}

\textbf{Real-world Performance.} As shown in Table~\ref{tab:real}, we evaluate RoTri-Diff across four real-world manipulation tasks using the number of successful trials (out of 5 trials per task).
In the ``Pick Tomato \& Banana” task, RoTri-Diff achieves 5 successful trials out of 5, demonstrating its ability to plan and execute two independent parallel grasps within a shared workspace while avoiding collisions.
In the ``Pick Plate” and ``Wash Plate” tasks involving complex sequential manipulation, RoTri-Diff achieves 3 out of 5 and 4 out of 5 successful trials, respectively, highlighting its capability to handle tasks with strict temporal dependencies and precise bimanual coordination. Notably, in the ``Pick Plate” task, one arm first tilts and stabilizes the plate before the other grasps the lifted side, imposing strict spatial constraints between the two arms and the object.
In the ``Lift Basket” task, RoTri-Diff achieves 4 successful trials out of 5, indicating its ability to maintain stable and synchronous spatial coordination between the two arms and the object under load.
Taken together, these results demonstrate that RoTri-Diff can robustly address the diverse complexities of real-world bimanual manipulation.

\section{Conclusion and Limitation}

We propose RoTri-Diff, a diffusion-based imitation learning framework for bimanual manipulation that explicitly models the robot–object triadic interaction (\textit{RoTri}) for the first time. The \textit{RoTri} representation encodes the relative 6D poses between the two end-effectors and the manipulated object, thereby capturing their triadic spatial relationship through continuous triangular geometric constraints. Built upon this representation, RoTri-Diff integrates robot keyposes, object pointflow, and \textit{RoTri} constraints within a hierarchical diffusion process to generate temporally and spatially stable bimanual actions. Extensive experiments show that RoTri-Diff not only outperforms state-of-the-art baselines across simulation tasks with different coordination modes, but also achieves stable performance in 4 real-world challenging tasks, paving the way toward human-level bimanual manipulation in diverse scenarios.

Despite its strengths, RoTri-Diff remains limited by its reliance on rigid-body assumptions and accurate 6D pose estimation, restricting generalization to deformable objects and unstructured environments. Future work will aim to establish a more general triadic interaction representation capable of handling deformable objects, supporting cross-embodiment transfer, and adapting to diverse bimanual platforms.


{\small
\bibliographystyle{IEEEtran}
\bibliography{short,references}

@inproceedings{ppi,
  title={Gripper Keypose and Object Pointflow as Interfaces for Bimanual Robotic Manipulation},
  author={Yang, Yuyin and Cai, Zetao and Tian, Yang and Zeng, Jia and Pang, Jiangmiao},
  booktitle=RSS,
  year={2025}
}

@inproceedings{peract2,
  title={Peract2: A perceiver actor framework for bimanual manipulation tasks},
  author={Markus Grotz and Mohit Shridhar and Tamim Asfour and Dieter Fox},
  booktitle = {Arxiv},
  year={2024}
}

@inproceedings{act,
  title={Learning fine-grained bimanual manipulation with low-cost hardware},
  author={Zhao, Tony Z and Kumar, Vikash and Levine, Sergey and Finn, Chelsea},
  booktitle=RSS,
  year={2023}
}

@inproceedings{dp3,
  title={3d diffusion policy},
  author={Ze, Yanjie and Zhang, Gu and Zhang, Kangning and Hu, Chenyuan and Wang, Muhan and Xu, Huazhe},
  booktitle=RSS,
  year={2024}
}

@inproceedings{3d-diffuser-actor,
  title={3d diffuser actor: Policy diffusion with 3d scene representations},
  author={Ke, Tsung-Wei and Gkanatsios, Nikolaos and Fragkiadaki, Katerina},
  booktitle=CORL,
  year={2024}
}

@inproceedings{any-bimanual,
  title={Anybimanual: Transferring unimanual policy for general bimanual manipulation},
  author={Lu, Guanxing and Yu, Tengbo and Deng, Haoyuan and Chen, Season Si and Tang, Yansong and Wang, Ziwei},
  booktitle=ICCV,
  year={2025}
}

@inproceedings{bikc,
  title={BiKC: Keypose-Conditioned Consistency Policy for Bimanual Robotic Manipulation},
  author={Yu, Dongjie and Xu, Hang and Chen, Yizhou and Ren, Yi and Pan, Jia},
  booktitle={Arxiv},
  year={2024}
}

@inproceedings{mba,
  title={Motion before action: Diffusing object motion as manipulation condition},
  author={Su, Yue and Zhan, Xinyu and Fang, Hongjie and Li, Yong-Lu and Lu, Cewu and Yang, Lixin},
  booktitle ={IEEE Robotics and Automation Letters},
  year={2025},
  publisher={IEEE}
}

@inproceedings{open-motion-planning,
  title={The open motion planning library},
  author={Sucan, Ioan A and Moll, Mark and Kavraki, Lydia E},
  booktitle ={IEEE Robotics \& Automation Magazine},
  volume={19},
  number={4},
  pages={72--82},
  year={2012},
  publisher={IEEE}
}

@inproceedings{ddpm,
  title={Denoising diffusion probabilistic models},
  author={Ho, Jonathan and Jain, Ajay and Abbeel, Pieter},
  booktitle=NEURIPS,
  year={2020}
}

@inproceedings{sdxl,
  title={SDXL: Improving Latent Diffusion Models for High-Resolution Image Synthesis},
  author={Podell, Dustin and English, Zion and Lacey, Kyle and Blattmann, Andreas and Dockhorn, Tim and M{\"u}ller, Jonas and Penna, Joe and Rombach, Robin},
  booktitle=ICLR,
  year={2024}
}

@inproceedings{rombach2022high,
  title={High-resolution image synthesis with latent diffusion models},
  author={Rombach, Robin and Blattmann, Andreas and Lorenz, Dominik and Esser, Patrick and Ommer, Bj{\"o}rn},
  booktitle=CVPR,
  year={2022}
}

@inproceedings{prolificdreamer,
  title={Prolificdreamer: High-fidelity and diverse text-to-3d generation with variational score distillation},
  author={Wang, Zhengyi and Lu, Cheng and Wang, Yikai and Bao, Fan and Li, Chongxuan and Su, Hang and Zhu, Jun},
   booktitle=NEURIPS,
  volume={36},
  year={2023}
}

@inproceedings{scalable,
  title={Scalable diffusion models with transformers},
  author={Peebles, William and Xie, Saining},
  booktitle=ICCV,
  year={2023}
}

@inproceedings{dp,
  title={Diffusion Policy: Visuomotor Policy Learning via Action Diffusion},
  author={Chi, Cheng and Feng, Siyuan and Du, Yilun and Xu, Zhenjia and Cousineau, Eric and Burchfiel, Benjamin and Song, Shuran},
  booktitle=RSS,
  year={2023}
}

@inproceedings{hdp,
  title={Hierarchical diffusion policy for kinematics-aware multi-task robotic manipulation},
  author={Ma, Xiao and Patidar, Sumit and Haughton, Iain and James, Stephen},
  booktitle=CVPR,
  year={2024}
}

@inproceedings{render-and-diffuser,
  title={Render and diffuse: Aligning image and action spaces for diffusion-based behaviour cloning},
  author={Vosylius, Vitalis and Seo, Younggyo and Uru{\c{c}}, Jafar and James, Stephen},
  booktitle=RSS,
  year={2024}
}

@inproceedings{diffusion,
  title={Diffusion policy policy optimization},
  author={Ren, Allen Z and Lidard, Justin and Ankile, Lars L and Simeonov, Anthony and Agrawal, Pulkit and Majumdar, Anirudha and Burchfiel, Benjamin and Dai, Hongkai and Simchowitz, Max},
  booktitle = {Arxiv},
  year={2024}
}

@inproceedings{sdp,
  title={Sparse Diffusion Policy: A Sparse, Reusable, and Flexible Policy for Robot Learning},
  author={Wang, Yixiao and Zhang, Yifei and Huo, Mingxiao and Tian, Thomas and Zhang, Xiang and Xie, Yichen and Xu, Chenfeng and Ji, Pengliang and Zhan, Wei and Ding, Mingyu and others},
  booktitle=CORL,
  year={2025}
}

@inproceedings{gravmad,
  title={GravMAD: Grounded Spatial Value Maps Guided Action Diffusion for Generalized 3D Manipulation},
  author={Chen, Yangtao and Chen, Zixuan and Yin, Junhui and Huo, Jing and Tian, Pinzhuo and Shi, Jieqi and Gao, Yang},
  booktitle=ICLR,
  year={2025}
}

@inproceedings{RDT-1B,
  title={RDT-1B: a Diffusion Foundation Model for Bimanual Manipulation},
  author={Liu, Songming and Wu, Lingxuan and Li, Bangguo and Tan, Hengkai and Chen, Huayu and Wang, Zhengyi and Xu, Ke and Su, Hang and Zhu, Jun},
  booktitle=ICLR,
  year={2025}
}

@inproceedings{D-CODA,
  title={D-CODA: Diffusion for Coordinated Dual-Arm Data Augmentation},
  author={Liu, I and Arthur, Chun and Chen, Jason and Sukhatme, Gaurav and Seita, Daniel},
  booktitle = {Arxiv},
  year={2025}
}

@inproceedings{kstar,
  title={Spatial-temporal graph diffusion policy with kinematic modeling for bimanual robotic manipulation},
  author={Lv, Qi and Li, Hao and Deng, Xiang and Shao, Rui and Li, Yinchuan and Hao, Jianye and Gao, Longxiang and Wang, Michael Yu and Nie, Liqiang},
  booktitle=CVPR,
  pages={17394--17404},
  year={2025}
}

@inproceedings{gdp3,
  title={Generalizable humanoid manipulation with improved 3d diffusion policies},
  author={Ze, Yanjie and Chen, Zixuan and Wang, Wenhao and Chen, Tianyi and He, Xialin and Yuan, Ying and Peng, Xue Bin and Wu, Jiajun},
  booktitle = {Arxiv},
  year={2024}
}

@inproceedings{dinov2,
  title={DINOv2: Learning Robust Visual Features without Supervision},
  author={Oquab, Maxime and Darcet, Timothée and Moutakanni, Theo and Vo, Huy V. and Szafraniec, Marc and Khalidov, Vasil and Fernandez, Pierre and Haziza, Daniel and Massa, Francisco and El-Nouby, Alaaeldin and Howes, Russell and Huang, Po-Yao and Xu, Hu and Sharma, Vasu and Li, Shang-Wen and Galuba, Wojciech and Rabbat, Mike and Assran, Mido and Ballas, Nicolas and Synnaeve, Gabriel and Misra, Ishan and Jegou, Herve and Mairal, Julien and Labatut, Patrick and Joulin, Armand and Bojanowski, Piotr},
  booktitle={Arxiv},
  year={2023}
}

@inproceedings{pointnet,
    author = {Charles R. Qi and Li Yi and Hao Su and Leonidas J. Guibas},
    title = {PointNet++: Deep Hierarchical Feature Learning on Point Sets in a Metric Space},
    booktitle = {Arxiv},
    year = 2017
}

@inproceedings{groundingdino,
  title={Grounding dino: Marrying dino with grounded pre-training for open-set object detection},
  author={Liu, Shilong and Zeng, Zhaoyang and Ren, Tianhe and Li, Feng and Zhang, Hao and Yang, Jie and Li, Chunyuan and Yang, Jianwei and Su, Hang and Zhu, Jun and others},
  booktitle={Arxiv},
  year={2023}
}

@inproceedings{sam,
  title={Segment Anything},
  author={Kirillov, Alexander and Mintun, Eric and Ravi, Nikhila and Mao, Hanzi and Rolland, Chloe and Gustafson, Laura and Xiao, Tete and Whitehead, Spencer and Berg, Alexander C. and Lo, Wan-Yen and Doll{\'a}r, Piotr and Girshick, Ross},
  booktitle={Arxiv},
  year={2023}
}

@inproceedings{foundationpose,
author        = {Bowen, Wen and Wei, Yang and Jan, Kautz and Stan, Birchfield},
title         = {{FoundationPose}: Unified 6D Pose Estimation and Tracking of Novel Objects},
booktitle     = CVPR,
year          = {2024},
}

@inproceedings{bundlesdf,
  title={Bundlesdf: Neural 6-dof tracking and 3d reconstruction of unknown objects},
  author={Wen, Bowen and Tremblay, Jonathan and Blukis, Valts and Tyree, Stephen and M{\"u}ller, Thomas and Evans, Alex and Fox, Dieter and Kautz, Jan and Birchfield, Stan},
  booktitle=CVPR,
  pages={606--617},
  year={2023}
}

@inproceedings{qattention,
    author = {Stephen James and Andrew J. Davison},
    title = {Q-attention: Enabling Efficient Learning for Vision-based Robotic Manipulation},
    booktitle = {IEEE Robotics and Automation Letters} ,
    year = 2022
}

@inproceedings{film,
    author = {Ethan Perez and Florian Strub and Harm de Vries and Vincent Dumoulin and Aaron Courville},
    title = {FiLM: Visual Reasoning with a General Conditioning Layer},
    booktitle = AAAI,
    year = 2017
}

@inproceedings{perceiverio,
  title={Perceiver IO: A General Architecture for Structured Inputs \& Outputs},
  author={Jaegle, Andrew and Borgeaud, Sebastian and Alayrac, Jean-Baptiste and Doersch, Carl and Ionescu, Catalin and Ding, David and Koppula, Skanda and Zoran, Daniel and Brock, Andrew and Shelhamer, Evan and others},
  booktitle=ICLR,
  year=2022
}

@inproceedings{umi,
	title={Universal Manipulation Interface: In-The-Wild Robot Teaching Without In-The-Wild Robots},
	author={Chi, Cheng and Xu, Zhenjia and Pan, Chuer and Cousineau, Eric and Burchfiel, Benjamin and Feng, Siyuan and Tedrake, Russ and Song, Shuran},
	booktitle=RSS,
	year={2024}
}

@inproceedings{dro,
  title={$\mathcal{D(R,O)}$ Grasp: A Unified Representation of Robot and Object Interaction for Cross-Embodiment Dexterous Grasping},
  author={Wei, Zhenyu and Xu, Zhixuan and Guo, Jingxiang and Hou, Yiwen and Gao, Chongkai and Cai, Zhehao and Luo, Jiayu and Shao, Lin},
  booktitle=ICRA,
  pages={4982--4988},
  year={2025},
  organization={IEEE}
}

@inproceedings{RobustDexGrasp,
      title={{RobustDexGrasp}: Robust Dexterous Grasping of General Objects},
      author={Zhang, Hui and Wu, Zijian and Huang, Linyi and Christen, Sammy and Song, Jie},
      booktitle=CoRL,
      year={2025}
    }

@inproceedings{curobo,
  title={Curobo: Parallelized collision-free robot motion generation},
  author={Sundaralingam, Balakumar and Hari, Siva Kumar Sastry and Fishman, Adam and Garrett, Caelan and Van Wyk, Karl and Blukis, Valts and Millane, Alexander and Oleynikova, Helen and Handa, Ankur and Ramos, Fabio and others},
  booktitle=ICRA,
  year={2023},
}

@inproceedings{stabilize,
  title={Stabilize to act: Learning to coordinate for bimanual manipulation},
  author={Grannen, Jennifer and Wu, Yilin and Vu, Brandon and Sadigh, Dorsa},
  booktitle=CORL,
  year={2023},
}

@inproceedings{openvla,
  title={OpenVLA: An Open-Source Vision-Language-Action Model},
  author={Kim, Moo Jin and Pertsch, Karl and Karamcheti, Siddharth and Xiao, Ted and Balakrishna, Ashwin and Nair, Suraj and Rafailov, Rafael and Foster, Ethan and Lam, Grace and Sanketi, Pannag and others},
  booktitle=CORL,
  year={2024}
}

@inproceedings{rt1,
  title={Rt-1: Robotics transformer for real-world control at scale},
  author={Brohan, Anthony and Brown, Noah and Carbajal, Justice and Chebotar, Yevgen and Dabis, Joseph and Finn, Chelsea and Gopalakrishnan, Keerthana and Hausman, Karol and Herzog, Alex and Hsu, Jasmine and others},
  booktitle={Arxiv},
  year={2022}
}

@inproceedings{palm-e,
  title={Palm-e: An embodied multimodal language model},
  author={Driess, Danny and Xia, Fei and Sajjadi, Mehdi SM and Lynch, Corey and Chowdhery, Aakanksha and Ichter, Brian and Wahid, Ayzaan and Tompson, Jonathan and Vuong, Quan and Yu, Tianhe and others},
  booktitle={Arxiv},
  year={2023}
}

@inproceedings{scar,
  title={Scar: Refining skill chaining for long-horizon robotic manipulation via dual regularization},
  author={Chen, Zixuan and Ji, Ze and Huo, Jing and Gao, Yang},
  booktitle=NEURIPS,
  year={2024}
}

@inproceedings{robotwin,
  title={Robotwin: Dual-arm robot benchmark with generative digital twins},
  author={Mu, Yao and Chen, Tianxing and Chen, Zanxin and Peng, Shijia and Lan, Zhiqian and Gao, Zeyu and Liang, Zhixuan and Yu, Qiaojun and Zou, Yude and Xu, Mingkun and others},
  booktitle=CVPR,
  year={2025}
}

@inproceedings{deco,
  title={DeCo: Task Decomposition and Skill Composition for Zero-Shot Generalization in Long-Horizon 3D Manipulation},
  author={Chen, Zixuan and Yin, Junhui and Chen, Yangtao and Huo, Jing and Tian, Pinzhuo and Shi, Jieqi and Hou, Yiwen and Li, Yinchuan and Gao, Yang},
  booktile={ArXiv},
  year={2025}
}

@inproceedings{3d_flowmatch_actor,
            author = {Gkanatsios, Nikolaos and Xu, Jiahe and Bronars, Matthew and Mousavian, Arsalan and Ke, Tsung-Wei and Fragkiadaki, Katerina},
            title = {3D FlowMatch Actor: Unified 3D Policy for Single- and Dual-Arm Manipulation},
            booktile={ArXiv},
            year = {2025}
        }

@STRING{CVPR = {CVPR}}

@STRING{ICCV = {ICCV}}

@STRING{NEURIPS = {NeurIPS}}

@STRING{ICLR = {ICLR}}

@STRING{ICRA = {ICRA}}

@STRING{CORL = {CoRL}}

@STRING{AAAI = {AAAI}}

@STRING{RSS = {RSS}}

@STRING{ARXIV = {arXiv.org}}

@STRING{CHI = {CHI}}
}

\end{document}